\documentclass[runningheads]{llncs}
\usepackage[T1]{fontenc}
\usepackage{graphicx,verbatim}
\usepackage[table,xcdraw]{xcolor}
\usepackage{color}
\usepackage{colortbl}
\usepackage{amsmath} 
\usepackage{amssymb}
\usepackage{newtxmath}
\usepackage{booktabs}
\usepackage{multirow}
\usepackage{hyperref}
\usepackage{pgf}
\begin{document}
\title{Revisiting 3D Medical Scribble Supervision: Benchmarking Beyond Cardiac Segmentation}
\titlerunning{Revisiting 3D Medical Scribble Supervision}
%
\begin{comment}  %% Removed for anonymized MICCAI 2025 submission
\author{First Author\inst{1}\orcidID{0000-1111-2222-3333} \and
Second Author\inst{2,3}\orcidID{1111-2222-3333-4444} \and
Third Author\inst{3}\orcidID{2222--3333-4444-5555}}
%
\authorrunning{F. Author et al.}
\institute{Princeton University, Princeton NJ 08544, USA \and
Springer Heidelberg, Tiergartenstr. 17, 69121 Heidelberg, Germany
\email{lncs@springer.com}\\
\url{http://www.springer.com/gp/computer-science/lncs} \and
ABC Institute, Rupert-Karls-University Heidelberg, Heidelberg, Germany\\
\email{\{abc,lncs\}@uni-heidelberg.de}}

\end{comment}

\author{
Karol Gotkowski\inst{1, 2} \and
Klaus H. Maier-Hein\inst{1, 2} \and
Fabian Isensee\inst{1, 2}
}
%index{Gotkowski, Karol}
%index{Maier-Hein, Klaus}
%index{Isensee, Fabian}
%
\authorrunning{Karol Gotkowski et al.}
\institute{
Division of Medical Image Computing, DKFZ, Heidelberg, Germany \\
\email{\{karol.gotkowski,f.isensee\}@dkfz.de} \and
Helmholtz Imaging, DKFZ, Heidelberg, Germany
}

\maketitle              % typeset the header of the contribution
\begin{abstract}
Scribble supervision has emerged as a promising approach for reducing annotation costs in medical 3D segmentation by leveraging sparse annotations instead of voxel-wise labels. While existing methods report strong performance, a closer analysis reveals that the majority of research is confined to the cardiac domain, predominantly using ACDC and MSCMR datasets. This over-specialization has resulted in severe overfitting, misleading claims of performance improvements, and a lack of generalization across broader segmentation tasks.
In this work, we formulate a set of key requirements for practical scribble supervision and introduce ScribbleBench, a comprehensive benchmark spanning over seven diverse medical imaging datasets, to systematically evaluate the fulfillment of these requirements. Consequently, we uncover a general failure of methods to generalize across tasks and that many widely used novelties degrade performance outside of the cardiac domain, whereas simpler overlooked approaches achieve superior generalization. Finally, we raise awareness for a strong yet overlooked baseline, nnU-Net coupled with a partial loss, which consistently outperforms specialized methods across a diverse range of tasks.
By identifying fundamental limitations in existing research and establishing a new benchmark-driven evaluation standard, this work aims to steer scribble supervision toward more practical, robust, and generalizable methodologies for medical image segmentation.

\keywords{scribble supervision  \and segmentation \and 3D \and medical.}
% Authors must provide keywords and are not allowed to remove this Keyword section.

\end{abstract}
\section{Introduction}
\begin{figure}
    \centering
    \includegraphics[width=1\textwidth]{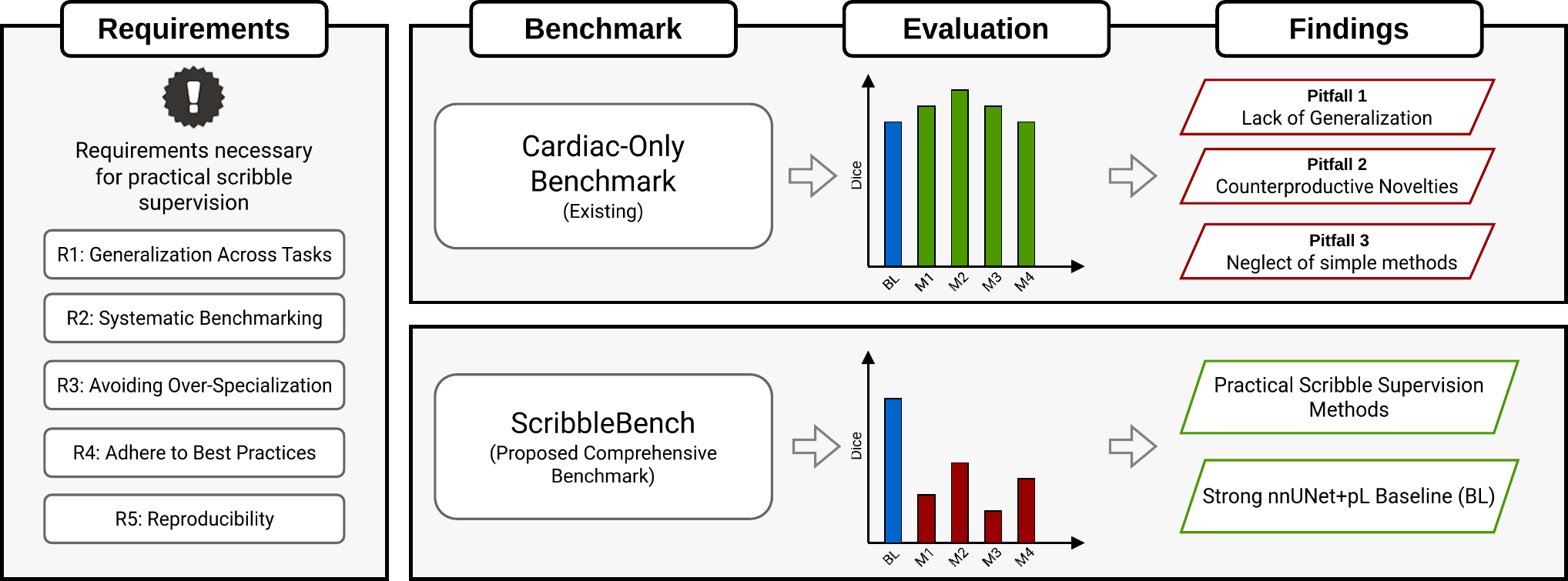}
    \caption{We define key requirements for practical scribble supervision and introduce a benchmark to assess existing methods, revealing major pitfalls in the process. Revisiting an overlooked baseline, we find it generalizes exceptionally well across new tasks.}
    \label{fig:figure1}
\end{figure}

Scribble supervision has emerged as a promising approach to significantly reduce annotation costs in 3D medical segmentation tasks. By using simple annotations in the form of scribbles, this approach allows for faster data labeling while maintaining practical applicability. In light of this promise, consistent improvements in training models from scribbles have been reported over the years, with current state-of-the-art methods \cite{li2024scribformer,zhang2022cyclemix,zhang2022shapepu,han2024dmsps} achieving Dice scores of 0.84 and higher, only a few Dice points lower than their densely supervised counterparts. \\
Numerous 3D scribble supervision methods \cite{han2023scribble,li2023scribblevc,zhang2022shapepu,zhang2022cyclemix,luo2022scribble,chen2022scribble2d5,valvano2021learning,lee2020scribble2label,ji2019scribble,can2018learning,luo2021word,han2024dmsps,zhang2023zscribbleseg,wong2024scribbleprompt} have been proposed that incorporate innovative strategies to improve segmentation performance, such as custom augmentations \cite{zhang2022cyclemix,zhang2022shapepu}, intricate regularized loss functions \cite{han2023scribble,li2023scribblevc,zhang2022cyclemix,zhang2022shapepu,chen2022scribble2d5,lee2020scribble2label,ji2019scribble,can2018learning}, pseudo-label learning \cite{li2023scribblevc,luo2022scribble,chen2022scribble2d5,lee2020scribble2label,ji2019scribble,can2018learning}, novel segmentation model architectures \cite{han2023scribble,li2023scribblevc,luo2022scribble,chen2022scribble2d5,valvano2021learning,ji2019scribble,can2018learning}, and multi-stage training \cite{han2024dmsps}. \\
Given the performance of current scribble supervision methods, it seems that the task of scribble supervision is mostly solved. However, fueled by the scarcity of publicly available expert scribbles, most research in this domain relies almost exclusively on the ACDC and MSCMR cardiac segmentation datasets for evaluation \cite{li2023scribblevc,li2024scribformer,luo2022scribble,zhang2022cyclemix,zhang2022shapepu,zhang2023zscribbleseg,zhou2023weakly,wang2024scribblevs}, henceforth referred to as the cardiac benchmark. Unknowingly, this specialization on the cardiac benchmark has steered scribble supervision research into a local optimum, succumbing to numerous validation pitfalls, limiting the potential of the proposed methods and confining scribble supervision to a niche research domain. Consequently, efforts must be made to refocus the field of scribble supervision toward generalizable methods with practical applicability (see Figure \ref{fig:figure1}). \\
Our study makes the following contributions:
\begin{enumerate}
  \item We propose a comprehensive set of requirements that scribble supervision methods should fulfill to be viable alternatives to dense annotations.
  \item We propose a comprehensive benchmark for the systematic evaluation of scribble supervision methods, ensuring alignment with the requirements.
  \item We reveal three fundamental validation pitfalls: 1) current methods severely overfit to datasets of the cardiac domain, 2) frequently include counterproductive methodological novelties that degrade performance, and 3) neglect simple methodologies that perform well and generalize better to new tasks.
  \item Based on these insights we raise awareness for a previously overlooked strong and robust baseline, which generalizes exceptionally well across all new tasks and thus marks the first out-of-the-box scribble supervision method that can reliably be used to train segmentation models on new datasets.
\end{enumerate}

\section{Requirements} \label{Requirements}
For scribble supervision to be practically viable, it must meet key requirements to ensure usability, reliability, and effectiveness across diverse medical segmentation tasks:
\begin{itemize}
  \item R1 - Generalization Across Tasks: Scribble supervision should work out-of-the-box across different anatomical structures and imaging modalities. Additionally, any newly introduced novelties should undergo systematic ablation to assess their true impact on generalization and performance.
  \item R2 - Systematic Benchmarking on Diverse Tasks: To accurately validate claims of generalization and prevent misleading results caused by dataset overfitting, evaluation on a comprehensive benchmark is necessary.
  \item R3 - Avoiding Over-Specialization: Methods should be designed to integrate seamlessly into various segmentation architectures to ease adaptation to new state-of-the-art segmentation models rather than being over-engineered frameworks for specific datasets.
  \item R4 - Maximizing Performance Through Established Practices: Methods should adhere to proven best practices, such as employing 3D segmentation architectures, which consistently outperform 2D approaches.
  \item R5 - Open-Source Implementation for Reproducibility: Publicly available code is crucial for fostering research transparency and enabling adoption in real-world applications.
\end{itemize}

\section{ScribbleBench: A Comprehensive Benchmark for Generalizable Scribble Supervision} \label{ScribbleBench}
\begin{figure}
    \centering
    \includegraphics[width=0.92\textwidth]{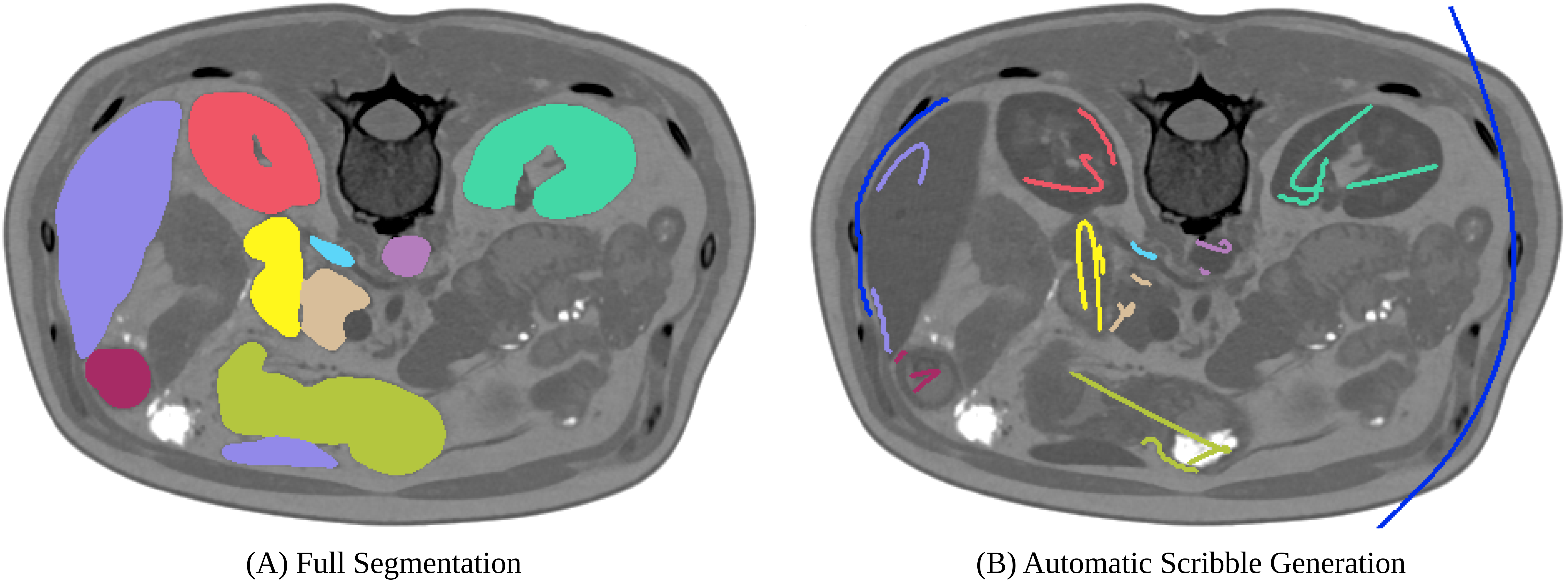}
    \caption{Examples of a dense segmentation (A) and our generated scribbles (B).}
    \label{fig:figure2}
\end{figure}

The lack of a comprehensive scribble supervision benchmark has so far hindered the development of robust scribble supervision methods as well as systematic comparisons between them. Previous approaches almost exclusively relied on the ACDC \cite{bernard2018deep,valvano2021learning} and MSCMR \cite{zhuang2016multivariate,zhuang2018multivariate} datasets, both representing the rather narrow task of cardiac segmentation in cine MRI. 
To address this, we created a new benchmark, ScribbleBench, for medical 3D scribble supervision with realistic scribbles, encompassing seven datasets of diverse segmentation tasks: LiTS \cite{bilic2023liver}, BraTS2020 \cite{menze2014multimodal}, AMOS2022 \cite{ji2022amos}, KiTS2023 \cite{KiTS23}, WORD \cite{luo2021word}, MSCMR \cite{zhuang2016multivariate,zhuang2018multivariate}, and ACDC \cite{bernard2018deep,valvano2021learning}. We automate scribble generation for these datasets using the existing dense reference segmentations. For each image slice, we generate two scribbles of different types for each class: interior scribbles mimicking human annotations placed randomly within the class interior as non-uniform rational B-Splines (NURBS), and border scribbles roughly delineating a small part of the class border with random slight offsets (see Figure \ref{fig:figure2}). This dual approach follows best practices to generate realistic scribbles as created by a human and ensures that both class interiors and boundaries are represented during training. 
Through the introduction of this benchmark, we aim to catalyze the development and comparison of new medical 3D scribble supervision methods.
\begin{table}[hb]
\centering
\caption{\textbf{Evaluation of Scribble Style Suitability:} Consistent Dice scores for ACDC and MSCMR across scribble styles validate the suitability of generated scribbles, while poor AMOS results indicate limited method generalizability.}
\resizebox{\columnwidth}{!}{%
\begin{tabular}{ccccccccc}
\toprule
\textbf{Method} & \multicolumn{2}{c}{\textbf{Expert Scribbles}} & \multicolumn{3}{c}{\textbf{SP Scribbles}} & \multicolumn{3}{c}{\textbf{Our Scribbles}} \\
\cmidrule(lr){2-3} \cmidrule(lr){4-6} \cmidrule(lr){7-9}
 & \textbf{ACDC} & \textbf{MSCMR} & \textbf{ACDC} & \textbf{MSCMR} & \textbf{AMOS} & \textbf{ACDC} & \textbf{MSCMR} & \textbf{AMOS} \\
\midrule
ShapePU           & \cellcolor[HTML]{c1ec94} 0.850 & \cellcolor[HTML]{c4ec95} 0.844 & \cellcolor[HTML]{d6ef96} 0.812 & \cellcolor[HTML]{ffbe9b} 0.475 & \cellcolor[HTML]{ffc09b} 0.484 & \cellcolor[HTML]{e9f198} 0.777 & \cellcolor[HTML]{ffc29b} 0.491 & \cellcolor[HTML]{ffa79b} 0.360\\
ScribFormer       & \cellcolor[HTML]{b0e993} 0.881 & \cellcolor[HTML]{c6ed95} 0.840 & \cellcolor[HTML]{ade993} 0.886 & \cellcolor[HTML]{f6f39a} 0.753 & \cellcolor[HTML]{ffc19b} 0.488 & \cellcolor[HTML]{b3ea93} 0.875 & \cellcolor[HTML]{d4ef96} 0.814 & \cellcolor[HTML]{ffaa9b} 0.375\\
CycleMix          & \cellcolor[HTML]{aee993} 0.884 & \cellcolor[HTML]{baeb94} 0.863 & \cellcolor[HTML]{a3e892} 0.905 & \cellcolor[HTML]{beeb94} 0.854 & \cellcolor[HTML]{ffce9b} 0.552 & \cellcolor[HTML]{a9e892} 0.894 & \cellcolor[HTML]{bfeb94} 0.854 & \cellcolor[HTML]{ffc19b} 0.486\\
DMSPS             & \cellcolor[HTML]{abe992} 0.891 & \cellcolor[HTML]{b3ea93} 0.874 & \cellcolor[HTML]{a3e892} 0.905 & \cellcolor[HTML]{c3ec95} 0.847 & \cellcolor[HTML]{ffd89b} 0.601 & \cellcolor[HTML]{ade992} 0.887 & \cellcolor[HTML]{baeb94} 0.862 & \cellcolor[HTML]{ffd69b} 0.592\\
\midrule
nnUNet (dense)    & \cellcolor[HTML]{d3d3d3} 0.924 & \cellcolor[HTML]{d3d3d3} 0.906 & \cellcolor[HTML]{d3d3d3} 0.924 & \cellcolor[HTML]{d3d3d3} 0.906 & \cellcolor[HTML]{d3d3d3} 0.860 & \cellcolor[HTML]{d3d3d3} 0.924 & \cellcolor[HTML]{d3d3d3} 0.906 & \cellcolor[HTML]{d3d3d3} 0.860\\
\bottomrule
\end{tabular}%
}
\label{table:table1}
\end{table}\\
In an initial analysis, we validated the suitability of ScribbleBench for benchmarking scribble supervision using ShapePU \cite{zhang2022shapepu}, CycleMix \cite{zhang2022cyclemix}, ScribFormer \cite{li2024scribformer}, and DMSPS \cite{han2024dmsps} alongside a densely supervised nnU-Net \cite{isensee2021nnu} with three annotation types—expert scribbles, our generated scribbles, and ScribblePrompt (SP) scribbles \cite{wong2024scribbleprompt}—across ACDC, MSCMR, and AMOS2022. Expert scribbles were available only for ACDC and MSCMR. \\
Results in Table \ref{table:table1} confirm that, for the cardiac benchmark, most methods performed consistently across different scribble styles. This demonstrates that our generated scribbles are realistic and suitable for scribble supervision. Interestingly, the performance of ShapePU decreases when moving from expert scribbles to generated scribbles. Given that this finding is highly consistent between two generated scribble styles indicates that ShapePU overfits to the scribble style found in the cardiac benchmark, and is not an artifact of our generated scribbles. Notably, all methods exhibited a significant performance drop on AMOS2022 compared to the densely supervised nnU-Net, again consistent between ScribblePrompt and our scribbles. Note that performance differences between scribble methods are related to ScribblePrompt generating 71\% more annotated voxels than our method. The findings on AMOS2022 indicate that existing methods might be highly sensitive to dataset shifts, which requires further investigation.

\section{Assessing the State of Scribble Supervision}
Motivated by the initial findings in section \ref{ScribbleBench}, we evaluated the scribble supervision methods with the full ScribbleBench benchmark (see Table \ref{table:table2}) using ShapePU \cite{zhang2022shapepu}, CycleMix \cite{zhang2022cyclemix}, ScribFormer \cite{li2024scribformer}, and DMSPS \cite{han2024dmsps}, revealing three key validation pitfalls that we identify as the primary obstacles in this field.

\begin{table}[h]
\centering
\caption{\textbf{Assessing the State of Scribble Supervision:} Evaluation of scribble supervision methods on the cardiac benchmark and ScribbleBench using Dice. Specialized methods perform well on the cardiac benchmark, but struggle on ScribbleBenchs diverse tasks, while simple methods such as DenseCRF and WORD generalize better.}
\resizebox{\columnwidth}{!}{
\begin{tabular}{c||cc||cccccccc}
\toprule
\multirow{2}{*}{Method} & \multicolumn{2}{c|}{Cardiac Bench.} & \multicolumn{8}{c}{ScribbleBench} \\
\cmidrule(lr){2-3} \cmidrule(lr){4-11}
 & ACDC & MSCMR & ACDC & MSCMR & WORD & LiTS & BraTS & AMOS & KiTS & Mean \\
\midrule
ShapePU & \cellcolor[HTML]{c2ec95} 0.850 & \cellcolor[HTML]{d2ee96} 0.844 & \cellcolor[HTML]{ffc79b} 0.777 & \cellcolor[HTML]{ffa79b} 0.491 & \cellcolor[HTML]{ffa79b} 0.465 & \cellcolor[HTML]{ffa79b} 0.234 & \cellcolor[HTML]{ffa79b} 0.057 & \cellcolor[HTML]{ffa79b} 0.360 & \cellcolor[HTML]{ffa79b} 0.195 & \cellcolor[HTML]{ffa79b} 0.369\\
ScribFormer & \cellcolor[HTML]{aae992} 0.881 & \cellcolor[HTML]{d6ef97} 0.840 & \cellcolor[HTML]{d3ee96} 0.875 & \cellcolor[HTML]{f1f399} 0.814 & \cellcolor[HTML]{ffc09b} 0.554 & \cellcolor[HTML]{ffe09b} 0.524 & \cellcolor[HTML]{ffd79b} 0.267 & \cellcolor[HTML]{ffaa9b} 0.375 & \cellcolor[HTML]{ffd19b} 0.424 & \cellcolor[HTML]{ffd59b} 0.548\\
CycleMix & \cellcolor[HTML]{a7e892} 0.884 & \cellcolor[HTML]{beeb94} 0.863 & \cellcolor[HTML]{a5e892} \textbf{0.894} & \cellcolor[HTML]{c6ec95} 0.854 & \cellcolor[HTML]{ffcb9b} 0.591 & \cellcolor[HTML]{ffe79b} 0.559 & \cellcolor[HTML]{ffa79b} 0.054 & \cellcolor[HTML]{ffc69b} 0.486 & \cellcolor[HTML]{ffda9b} 0.474 & \cellcolor[HTML]{ffd79b} 0.559\\
DMSPS & \cellcolor[HTML]{a3e892} \textbf{0.891} & \cellcolor[HTML]{b3ea93} \textbf{0.874} & \cellcolor[HTML]{b6ea93} 0.887 & \cellcolor[HTML]{bceb94} \textbf{0.862} & \cellcolor[HTML]{a3e892} \textbf{0.843} & \cellcolor[HTML]{e8f198} 0.660 & \cellcolor[HTML]{a3e892} \textbf{0.687} & \cellcolor[HTML]{ffe19b} 0.592 & \cellcolor[HTML]{ffc39b} 0.347 & \cellcolor[HTML]{eef299} 0.697\\
\midrule
nnUNet+DenseCRF & \cellcolor[HTML]{ffeb9b} 0.741 & \cellcolor[HTML]{ffd29b} 0.732 & \cellcolor[HTML]{abe992} 0.891 & \cellcolor[HTML]{d5ef96} 0.840 & \cellcolor[HTML]{aee993} 0.829 & \cellcolor[HTML]{d5ef96} 0.686 & \cellcolor[HTML]{ffef9b} 0.370 & \cellcolor[HTML]{beeb94} 0.790 & \cellcolor[HTML]{bfec94} 0.759 & \cellcolor[HTML]{d4ee96} 0.738\\
nnUNet+WORD & \cellcolor[HTML]{ffa79b} 0.519 & \cellcolor[HTML]{ffa79b} 0.645 & \cellcolor[HTML]{ffa79b} 0.720 & \cellcolor[HTML]{ffe29b} 0.729 & \cellcolor[HTML]{cbed95} 0.795 & \cellcolor[HTML]{a7e892} \textbf{0.750} & \cellcolor[HTML]{afe993} 0.648 & \cellcolor[HTML]{a5e892} \textbf{0.835} & \cellcolor[HTML]{aae992} \textbf{0.806} & \cellcolor[HTML]{c9ed95} \textbf{0.755}\\
\midrule
nnUNet (dense superv.) & \cellcolor[HTML]{d3d3d3} 0.924 & \cellcolor[HTML]{d3d3d3} 0.906 & \cellcolor[HTML]{d3d3d3} 0.924 & \cellcolor[HTML]{d3d3d3} 0.906 & \cellcolor[HTML]{d3d3d3} 0.861 & \cellcolor[HTML]{d3d3d3} 0.770 & \cellcolor[HTML]{d3d3d3} 0.827 & \cellcolor[HTML]{d3d3d3} 0.860 & \cellcolor[HTML]{d3d3d3} 0.846 & \cellcolor[HTML]{d3d3d3} 0.856\\
\bottomrule
\end{tabular}
}
\label{table:table2}
\end{table}

\subsection{P1: Limited evaluation hides a lack of generalization}
When methods are evaluated only on the cardiac benchmark, their generalizability to other tasks remains unverified, limiting their broader applicability. While all methods achieve high mean Dice scores ($\ge$0.84) 
\begin{figure}[h]
    \centering
    \includegraphics[width=0.75\textwidth]{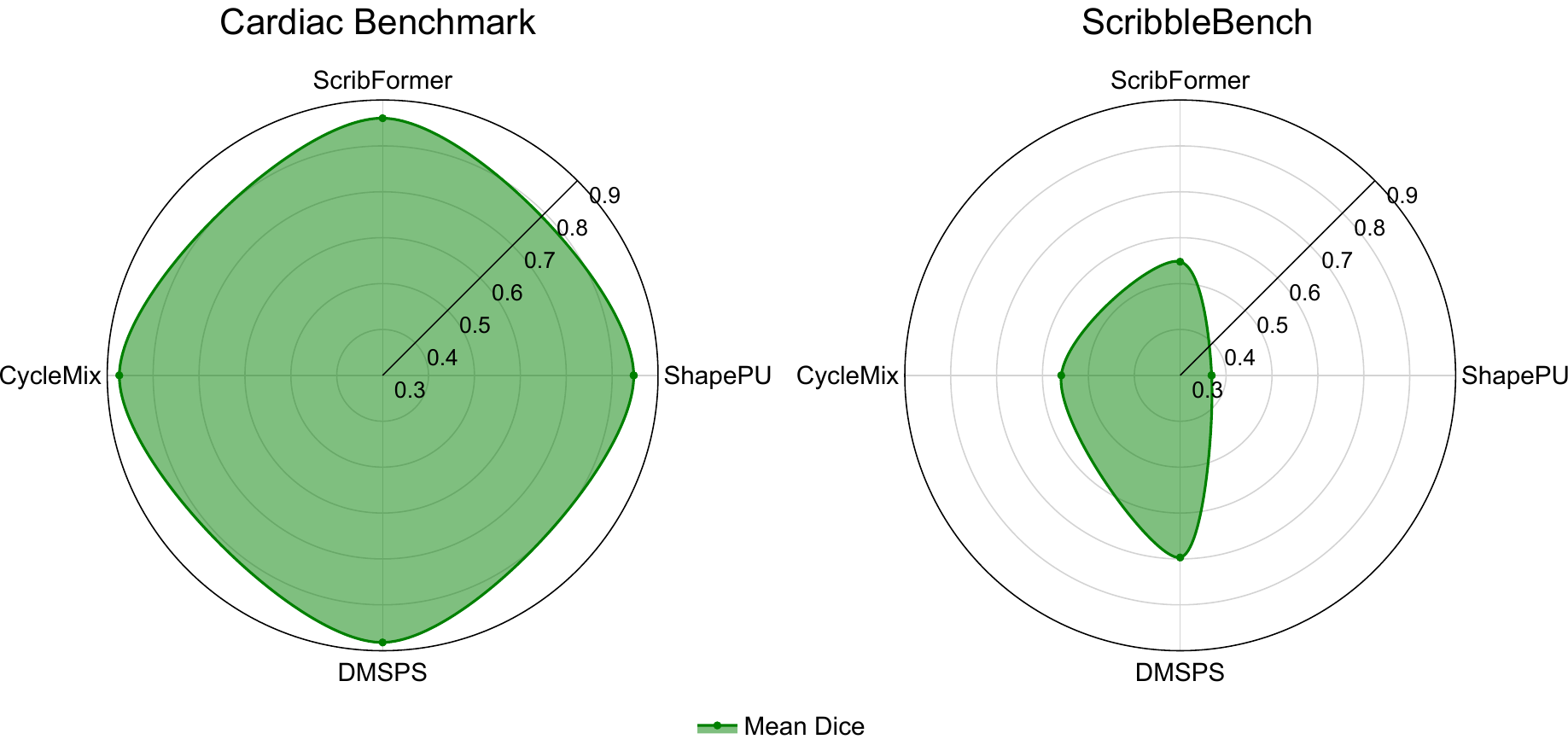}
    \caption{\textbf{Evaluating Generalization Across Benchmarks:} Strong performance on limited benchmarks sharply contrasts with significant declines on diverse tasks, highlighting generalization challenges.}
    \label{fig:figure3}
\end{figure}
on the cardiac benchmark (Figure \ref{fig:figure3}, left), creating the impression that scribble supervision is a solved problem, their performance drops markedly on ScribbleBench (Figure \ref{fig:figure3}, right). The significant Dice score reductions for ShapePU, CycleMix, ScribFormer, and DMSPS relative to the densely supervised upper bound, reveal that these methods lack generalization beyond cardiac segmentation (R1). This emphasizes the need for a diverse benchmark to accurately evaluate scribble supervision methods (R2).

\subsection{P2: Superficial novelties lead to performance degradation}
\begin{figure}
    \centering
    \includegraphics[width=0.75\textwidth]{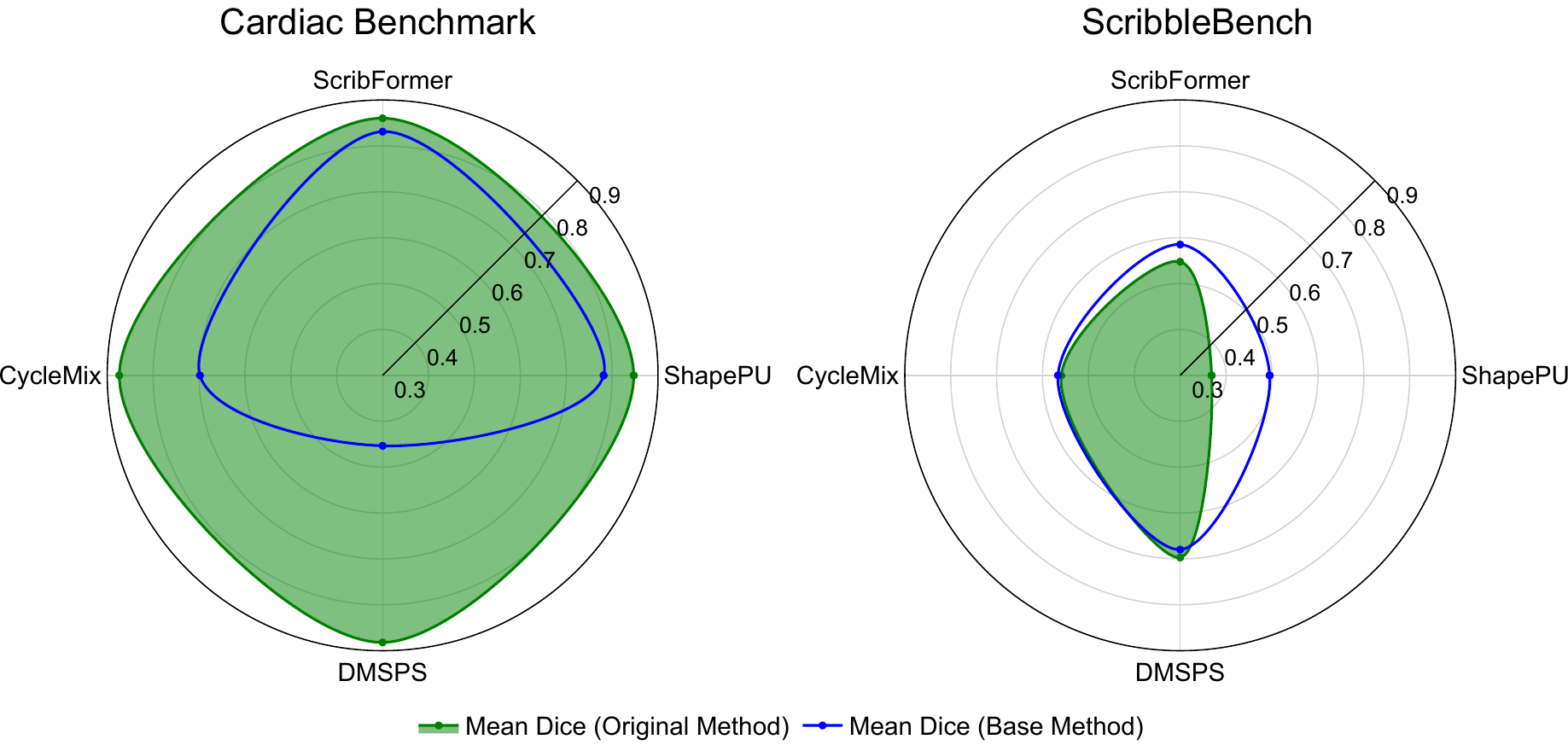}
    \caption{\textbf{Evaluating Novelty Impact Across Benchmarks:} While novelties improve performance on limited benchmarks, they lead to performance degradation relative to their respective base model when tested on a broader range of tasks. This hints at task overfitting and calls for a more comprehensive evaluation of methods.
    }
    \label{fig:figure4}
\end{figure}

Most scribble supervision methods are built upon a U-Net model and the partial Cross-Entropy loss (pCE) as a base for enabling scribble supervision. To enhance segmentation performance, researchers introduce a variety of additional techniques that greatly improve Dice performance by up to 0.429 on the cardiac benchmark (Figure \ref{fig:figure4}, left). However, when evaluated on ScribbleBench (Figure \ref{fig:figure4}, right) these novelties offer no benefit or, in most cases, are even detrimental when tested across a broader range of tasks as is the case with ShapePU, ScribFormer and CycleMix with considerable Dice reductions relative to the densely supervised upper bound. The respective proposed novelties result in substantial performance degradations and a marked reduction in generalization capability, highlighting the limitations of methods overly tailored to specific datasets (R2 \& R3).

\subsection{P3: Neglect of simple generalizing methods}
\begin{figure}
    \centering
    \includegraphics[width=0.75\textwidth]{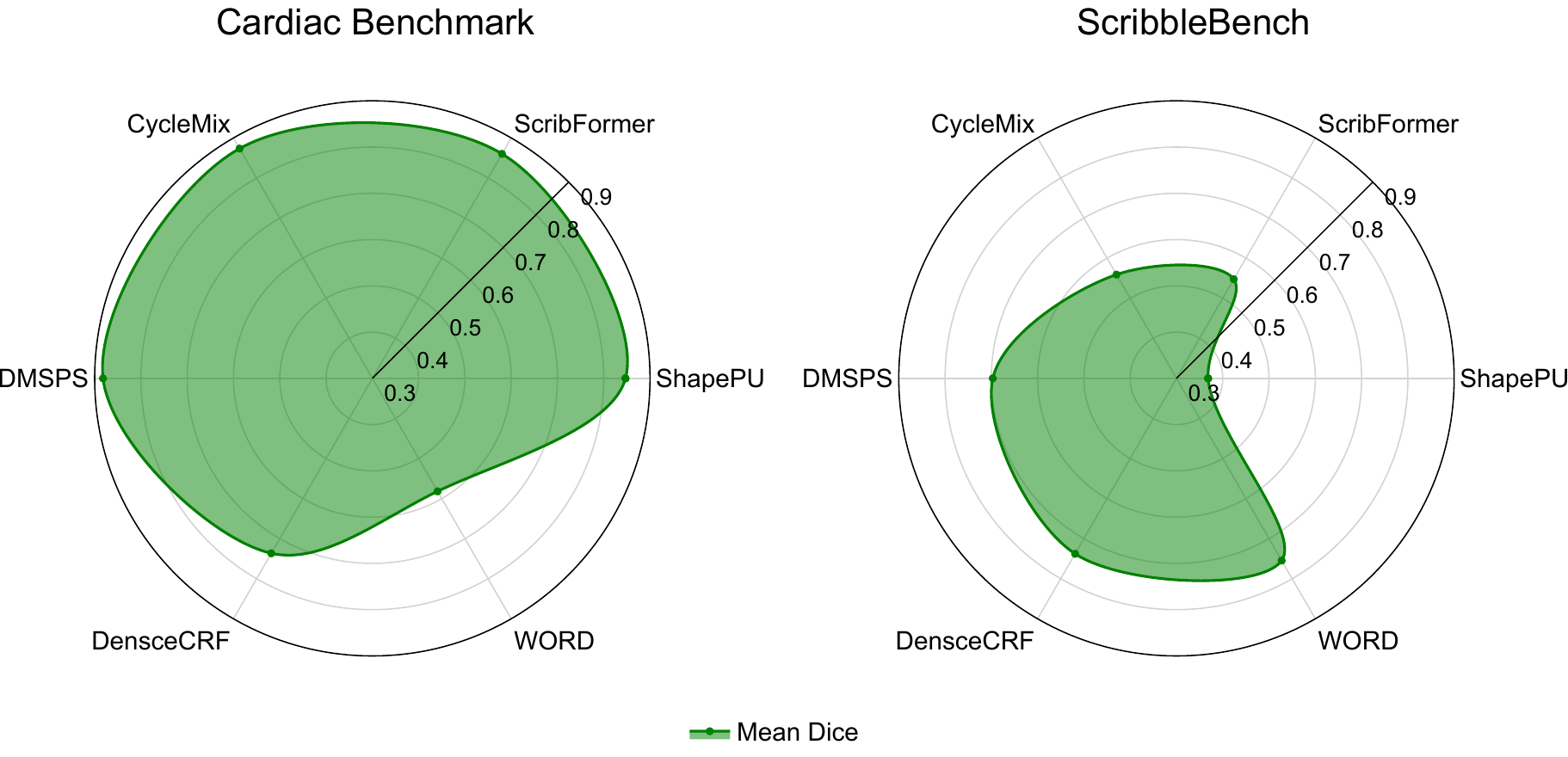}
    \caption{\textbf{Evaluating Simple Generalizing Methods:} Specialized methods outperform simpler alternatives on the cardiac benchmark, but generalization testing reveals that simple methods perform more consistently across diverse tasks.}
    \label{fig:figure5}
\end{figure}

The emphasis on specialized datasets, particularly the cardiac benchmark, has led researchers to overlook straightforward methods that generalize more effectively across diverse tasks (R3). Rather than relying on complex novelties, methods such as DenseCRF \cite{tang2018normalized} and WORD \cite{luo2021word} apply only minor modifications to the pCE loss, representing relatively simple yet impactful methodological changes. When evaluated on the cardiac benchmark (Figure \ref{fig:figure5}, left), these simpler methods appear to underperform relative to the more intricate, specialized approaches with mean Dice scores of 0.737 and 0.582, respectively. However, when assessed on our ScribbleBench (Figure \ref{fig:figure5}, right), they demonstrate superior generalization across multiple tasks with mean Dice scores of 0.738 and 0.755, respectively, while the specialized methods struggle. By focusing too heavily on highly specialized datasets, researchers risk disregarding simpler yet broadly effective approaches such as DenseCRF and WORD. 

\section{Establishing a Robust Baseline for Scribble Supervision}
The identified pitfalls highlight the crucial need for a rigorous evaluation on a diverse and comprehensive benchmark. Additionally, they reinforce the importance of adhering to the formulated requirements for effective scribble supervision methods. 
Building upon our insights into the generalization capabilities of simple loss modifications such as WORD and DenseCRF, we revisit the concept of the partial Cross Entropy loss (pCE), which only considers scribble-labeled voxels for the loss computation. This concept can be extended to other loss functions such as the Dice loss or combinations of losses as they are frequently used in state-of-the-art segmentation models. We denote the partial version of such losses simply as partial loss (pL). 
Given that nnU-Net \cite{isensee2021nnu} consistently achieves state-of-the-art segmentation performance \cite{isensee2024nnu} and generalizes effectively across datasets \cite{isensee2021nnu}, we adopt it as our base architecture. To add scribble supervision capabilities to nnU-Net we adapt its combined Cross-Entropy and Dice loss to its partial version (pL) and evaluate it on both the cardiac benchmark and ScribbleBench. Furthermore, we conduct ablations utilizing only pCE and the 2D architecture variant of nnU-Net.
The results, presented in Table \ref{table:table3} indicate that our baselines are slightly outperformed on the cardiac benchmark (see Table \ref{table:table2}). However, when evaluated on ScribbleBench, they exhibit exceptional generalization across all datasets with a mean of 0.813 for nnUNet+pL. Additionally, we observe a consistent improvement for our 3D baselines over their 2D counterparts (+0.061) and superior results for nnUNet+pL over nnUNet+pCE (+0.043). Notably, nnU-Net+pL surpasses all other specialized methods as well as WORD and DenseCRF (see Table \ref{table:table2}), often by a substantial margin. This suggests that nnU-Net+pL is a robust and practical approach for scribble supervision and serves as a strong baseline for future research in the field.

\begin{table}[t]
\centering
\caption{\textbf{Establishing a Strong Generalizable Baseline:} Evaluation of our simple method on the cardiac benchmark and ScribbleBench using Dice (same evaluation setting as in Table \ref{table:table2}). We observe that nnUNet+pL demonstrates exceptional generalization capabilities on ScribbleBench, constituting a strong baseline.}
\resizebox{\columnwidth}{!}{
\begin{tabular}{c||cc||cccccccc}
\toprule
\multirow{2}{*}{Method} & \multicolumn{2}{c|}{Cardiac Bench.} & \multicolumn{8}{c}{ScribbleBench} \\
\cmidrule(lr){2-3} \cmidrule(lr){4-11}
 & ACDC & MSCMR & ACDC & MSCMR & WORD & LiTS & BraTS & AMOS & KiTS & Mean \\
\midrule
nnUNet+pCE (2D) & \cellcolor[HTML]{ffc69b} 0.620 & \cellcolor[HTML]{ffdd9b} 0.753 & \cellcolor[HTML]{ffee9b} 0.846 & \cellcolor[HTML]{fff19b} 0.789 & \cellcolor[HTML]{afe993} 0.828 & \cellcolor[HTML]{d9ef97} 0.680 & \cellcolor[HTML]{ffe49b} 0.324 & \cellcolor[HTML]{bceb94} 0.793 & \cellcolor[HTML]{baeb94} 0.769 & \cellcolor[HTML]{e0f098} 0.718\\
nnUNet+pL (2D) & \cellcolor[HTML]{ffd09b} 0.653 & \cellcolor[HTML]{ffbe9b} 0.691 & \cellcolor[HTML]{a4e892} 0.894 & \cellcolor[HTML]{b9eb94} 0.866 & \cellcolor[HTML]{abe992} \textbf{0.833} & \cellcolor[HTML]{d2ee96} 0.691 & \cellcolor[HTML]{eff299} 0.442 & \cellcolor[HTML]{b9eb94} 0.798 & \cellcolor[HTML]{c7ed95} 0.740 & \cellcolor[HTML]{caed95} 0.752\\
nnUNet+pCE & \cellcolor[HTML]{d3ee96} 0.828 & \cellcolor[HTML]{a3e892} \textbf{0.890} & \cellcolor[HTML]{b6ea93} 0.887 & \cellcolor[HTML]{a4e892} 0.885 & \cellcolor[HTML]{c8ed95} 0.799 & \cellcolor[HTML]{a3e892} \textbf{0.756} & \cellcolor[HTML]{f7f39a} 0.418 & \cellcolor[HTML]{a4e892} 0.837 & \cellcolor[HTML]{a7e892} 0.812 & \cellcolor[HTML]{beeb94} 0.770\\
nnUNet+pL & \cellcolor[HTML]{c1ec94} \textbf{0.852} & \cellcolor[HTML]{b5ea93} 0.872 & \cellcolor[HTML]{a3e892} \textbf{0.895} & \cellcolor[HTML]{a3e892} \textbf{0.886} & \cellcolor[HTML]{bbeb94} 0.814 & \cellcolor[HTML]{a4e892} 0.753 & \cellcolor[HTML]{a5e892} \textbf{0.680} & \cellcolor[HTML]{a3e892} \textbf{0.840} & \cellcolor[HTML]{a3e892} \textbf{0.823} & \cellcolor[HTML]{a3e892} \textbf{0.813}\\
\midrule
nnUNet (dense superv.) & \cellcolor[HTML]{d3d3d3} 0.924 & \cellcolor[HTML]{d3d3d3} 0.906 & \cellcolor[HTML]{d3d3d3} 0.924 & \cellcolor[HTML]{d3d3d3} 0.906 & \cellcolor[HTML]{d3d3d3} 0.861 & \cellcolor[HTML]{d3d3d3} 0.770 & \cellcolor[HTML]{d3d3d3} 0.827 & \cellcolor[HTML]{d3d3d3} 0.860 & \cellcolor[HTML]{d3d3d3} 0.846 & \cellcolor[HTML]{d3d3d3} 0.856\\
\bottomrule
\end{tabular}
}
\label{table:table3}
\end{table}

\section{Conclusion}
Scribble supervision offers a cost-effective alternative to dense annotation in medical image segmentation. However, research has mainly focused on the cardiac benchmark, leaving claims of generalizability unverified. To address this, we defined key requirements for practical scribble supervision, emphasizing generalization, systematic benchmarking, and reproducibility. We introduced ScribbleBench, a benchmark covering seven diverse datasets, enabling comprehensive evaluation. Our findings expose key pitfalls in existing methods, showing that many fail to generalize and frequently even degrade performance beyond the cardiac domain. Lastly, we formulated a new baseline for scribble supervision: nnU-Net with a partial loss. By benchmarking its generalization with ScribbleBench (R1), systematically ablating its components (R2), using a generic loss (R3), building on the current state-of-the-art (R4), and releasing the code (R5), we demonstrate how adhering to the requirements enables researchers to identify methods that consistently outperform complex, specialized methods. We seek to realign scribble supervision research toward more generalizable, practical approaches by leveraging ScribbleBench and our robust baseline to enable out-of-the-box applicability and guide the field of scribble supervision to become a viable alternative to dense annotations. ScribbleBench, our scribble generation code and nnU-Net+pL will be released upon acceptance.

\bibliographystyle{splncs04}
\bibliography{mybibliography}




% LNCS Homepage, \url{http://www.springer.com/lncs}, last accessed 2023/10/25
% \end{thebibliography}
\end{document}